\documentclass{article}

\usepackage[nonatbib,preprint]{neurips_2021}

\usepackage[utf8]{inputenc} %
\usepackage[T1]{fontenc}    %
\usepackage{hyperref}       %
\usepackage{url}            %
\usepackage{booktabs}       %
\usepackage{amsfonts}       %
\usepackage{nicefrac}       %
\usepackage{microtype}      %
\usepackage{xcolor}         %
\usepackage{biblatex}
\usepackage{graphicx}
\usepackage{amsmath}

\title{Modeling human intention inference in continuous 3D domains by inverse planning and body kinematics}

\author{%
  Yingdong Qian\thanks{Department of Computer Science, University of California, Los Angeles} \\
   \texttt{yd@ucla.edu} \\
   \And
   Marta Kryven\thanks{Department of Brain and Cognitive Sciences, 
        Massachusetts Institute of Technology} \\
   \texttt{mkryven@mit.edu} \\
   \And
   Tao Gao\thanks{Department of Statistics and Communication, 
University of California, Los Angeles} \\
   \texttt{tao.gao@stat.ucla.edu} \\
   \And
   Hanbyul Joo\thanks{The Robotics Institute, Carnegie Mellon University} \\
   \texttt{hanbyulj@cs.cmu.edu} \\
   \And
   Josh Tenenbaum\thanks{Department of Brain and Cognitive Sciences, 
        Massachusetts Institute of Technology} \\
   \texttt{jbt@mit.edu} \\
}

\begin{document}

\maketitle
\begin{abstract}
How to build AI that understands human intentions, and uses this knowledge to collaborate with people? We describe a computational framework for evaluating models of goal inference in the domain of 3D motor actions, which receives as input the 3D coordinates of an agent’s body, and of possible targets, to produce a continuously updated inference of the intended target. We evaluate our framework in three behavioural experiments using a novel Target Reaching Task, in which human observers infer intentions of actors reaching for targets among distracts.
We describe Generative Body Kinematics model, which predicts human intention inference in this domain using Bayesian inverse planning %
and inverse body kinematics. We compare our model to three heuristics, which formalize the principle of least effort using simple assumptions about the actor’s constraints, without the use of inverse planning. Despite being more computationally costly, the Generative Body Kinematics model outperforms the heuristics in certain scenarios, such as environments with obstacles, and at the beginning of reaching actions while the actor is relatively far from the intended target. The heuristics make increasingly accurate predictions during later stages of reaching actions, such as, when the intended target is close, and can be inferred by extrapolating the wrist trajectory. Our results identify contexts in which inverse body kinematics is useful for intention inference. We show that human observers indeed rely on inverse body kinematics in such scenarios, suggesting that modeling body kinematic can improve performance of inference algorithms.
\end{abstract}

\section{Introduction}

Intention inference is one of the central questions in artificial intelligence (AI), critical to efficient human-AI collaboration, autonomous driving and automation.
In humans, intention inference has been perfected by evolutionary pressures, and emerges during the first six months of life \cite{woodward1998infants}. Six to 12 month old infants are already sensitive to the costs of other agents’ actions, showing surprise when someone takes a longer route to a goal when a shorter route is available \cite{liu2017six, gergely2003teleological}. 
The efficiency of such inferences relies on two core cognitive abilities. First, humans understand the actions of others in terms of optimized utilities -- the \textit{rationality principle}~\cite{dennett1989intentional, jara2016naive} or principle of least effort \cite{zipf1949human}. Second, humans understand physical interactions in terms of \textit{intuitive physics} --  mental approximations of physics in daily interactions~\cite{ullman2017mind}. 

The rationality principle states that agents minimize effort to achieve their goals, but does not demand that the agent's utilities are computed exactly \cite{jara2016naive}.  Assuming rationality has proved useful to quantitative modeling of human intention inference in 2D domains \cite{baker2009action, jara2016naive, kryven2021outcome,holtzen2016inferring}.
At the same time, studies show that humans use a variety of expected utility approximations in planning tasks~\cite{kryven2021plans,jain2019measuring,huys2015interplay}. 
Understanding which utility functions do people use, and when, is an active area of research, with a potential impact on building AI aligned with human values~\cite{hadfield2016cooperative,evans2016learning}.

Many previous studies on inferring human intentions used inverse reinforcement learning~\cite{ziebart2008maximum,hadfield2016cooperative,brown2019deep,ramachandran2007bayesian}, as well as deep-learning for inferring intentions of stereotyped behaviours~\cite{xing2020ensemble}.%
However, here we are interested in general-case intention inference in continuous motor domains, where human actions could be mediated by unpredictable environment layout, and include deviations from direct trajectories -- such as, for example, to recover balance, or avoid obstacles. For example, consider inferences of intentions of a soccer player, who is watching an opponent; or consider a driver watching a child who may be likely to run across the street.
In such scenarios intention attribution may require reasoning about physics~\cite{ullman2017mind}, motion trajectories~\cite{huang2020motion} and kinematics~\cite{vaziri2017predicting}.

\begin{figure*}[t]
  \centering
  \includegraphics[width=1.0\textwidth]{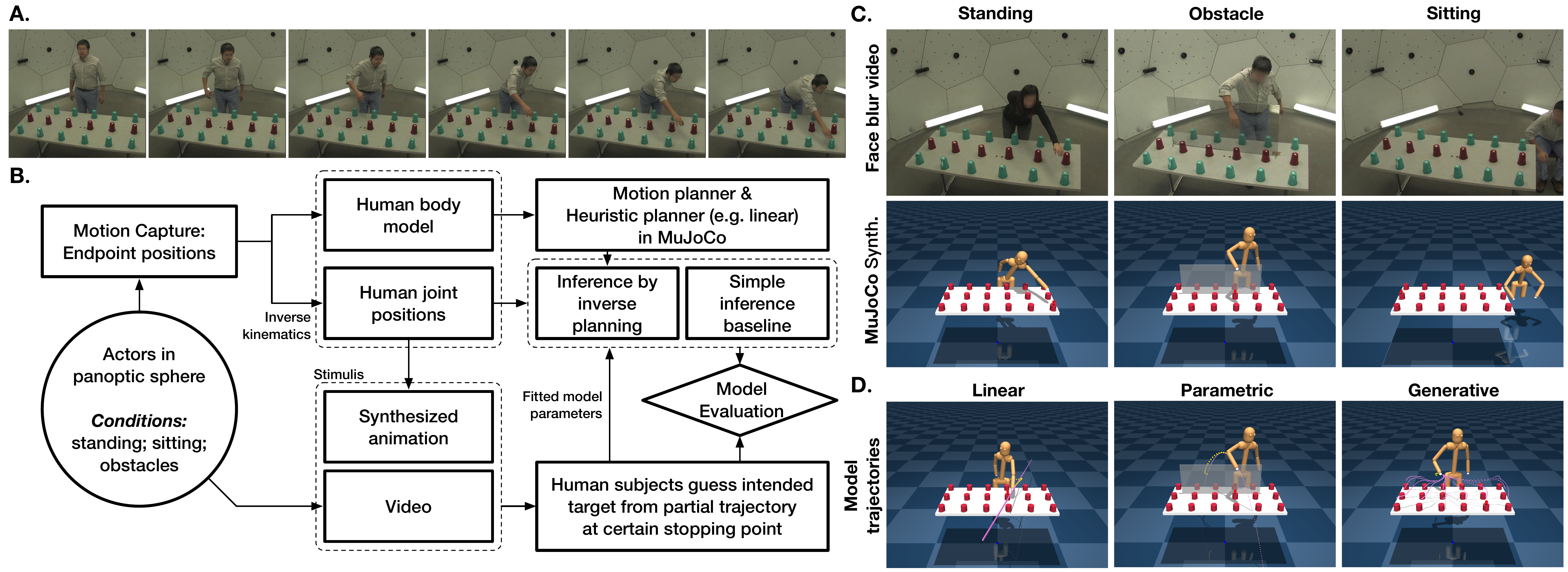}
  \caption{ \textbf{A.} A sequence of frames sampled from a stimulus in the Standing video condition. \textbf{B.} A diagram explaining our computational framework, stimuli generation procedure, and behavioural experiment. \textbf{(C)} Examples of environments shown in the six experimental conditions. In the obstacle condition, a transparent Plexiglas obstacle was placed in the middle of the table. The actor has to reach over the obstacle to reach objects on the outer side of the table. \textbf{(D)} Reaching trajectories predicted by the three models. LinH extrapolates a straight line from the history of observed wrist end-points, ParamH extrapolates a parametric curve in 3D space, and BodyGen simulates the most efficient movement of the skeletal figure to reach the 18 targets given the current position, acceleration and joint constraints. BodyGen complies with the laws of physics (e.g. Obstacles cannot be passed through) and follows the least effort principle. }
  \label{fig:1}
\end{figure*}

We hypothesized that in certain general-case scenarios humans use \textit{inverse kinematics} -- or simulation of the movements of an actor's body to anticipate the actor's goal -- which we formalize by our Generative Body Kinematics model.
At the same time, because of the high computational costs of inference in  high-dimensional continuous motor domains~\cite{wang2013probabilistic}, we also consider three simpler heuristic alternatives, which formalize minimization of effort without inverse planning, with less computations: Linear Extrapolation, Parametric Curve Extrapolation and Distance Heuristics. 
We show that the our model outperforms the three heuristics in certain scenarios, suggesting that the performance of intention inference algorithms can be improved by simulating body kinematics. 
We also show that the heuristics predict human inferences increasingly well at later stages of reaching actions, as the actor is nearing the target, suggesting that humans likely minimize computational effort in contexts were simple extrapolations of trajectory can achieve accurate results. 
To our knowledge, our model is the first to combine a 3D physics engine with inverse planning in a computational framework of intention inference in a real-life physical environment.

\section{Target Reaching Task}

\textbf{Hypothesis.} 
To test whether humans use inverse kinematics for goal inference we define our task domain as follows. A human actor is tasked with picking up a certain target object among multiple distractors. The actor knows the location of the target, and may need to walk, lean, or reach over an obstacle to pick it up. Each trial starts with the actor in a neutral standing position, and ends when the actor's hand touches the target. 
Fig.\ref{fig:1}A shows several example frames from such a recording. %
Fig.\ref{fig:1}B illustrates the flow-chart of our experimental framework.

\textbf{Data recording.} The actors' actions were recorded on video, as well as in 3D, using Panoptic Sphere -- a multi-view camera system designed to re-construct human skeleton and motion in 3D~\cite{joo2015panoptic}. We recorded two individuals, a male and a female, who used both left and right hand to pick up the objects. The recordings were segmented into trials, and post-processed to blur out the actor's face, to ensure that observers base their inferences on body motion, without relying on gaze. 
The recordings of 3D body positions were used as input for inference, as well as to generate skeletal animation stimuli (see below). 
Since the Panoptic Sphere does not record the positions of hands, or fingers, all modeling analysis was based on the position of the wrists. %

\begin{figure*}[t]
\centering
    \includegraphics[width=0.95\textwidth]{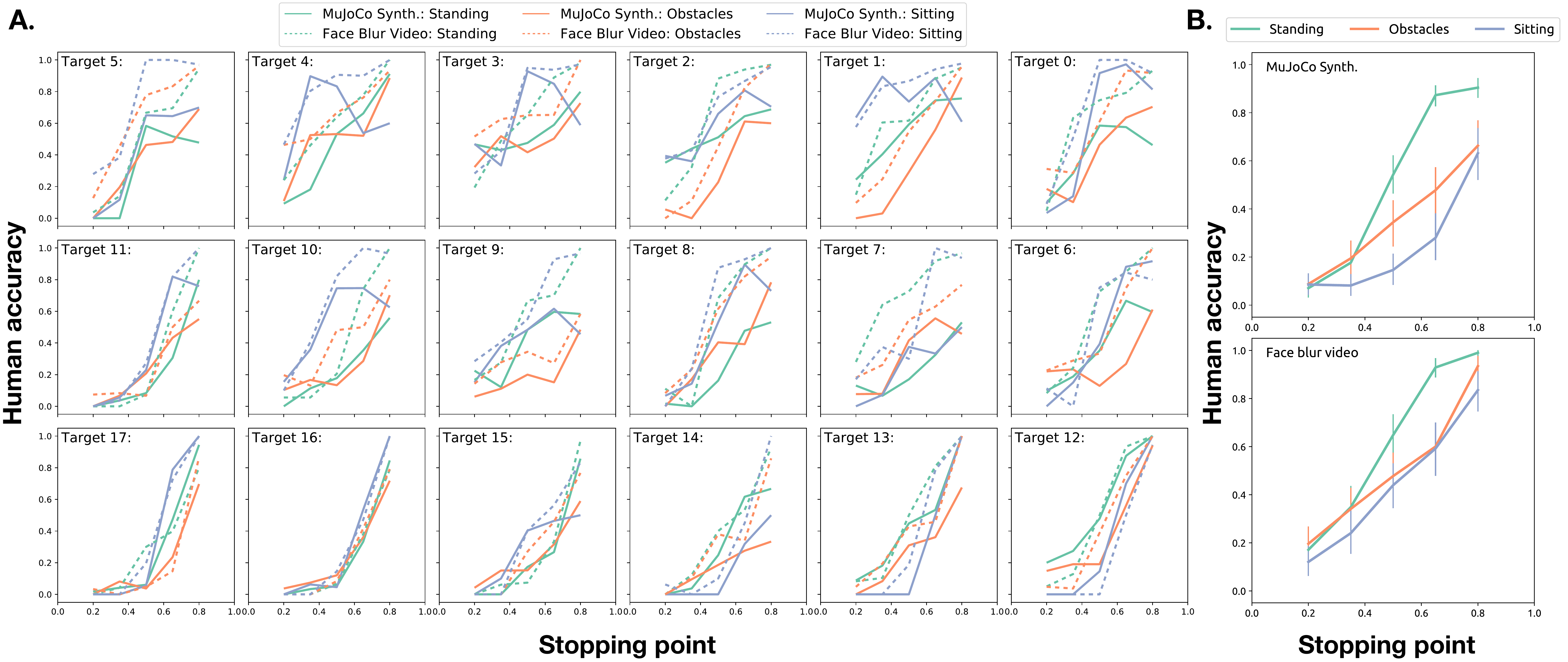}
    \caption{  Human performance across conditions. \textbf{(A)} Target detection accuracy in the six conditions. The grid layout corresponds to the object placement on the table in the experimental setup. Human accuracy differ among different targets. \textbf{(B)} Human performance in the six conditions aggregated across targets. The Standing conditions are the easiest, humans achieve a near-perfect accuracy after seeing $80\%$ of frames in both Face Blur and Synthetic stimuli. The Sitting conditions are the hardest. Due to 3D noise, accuracy is overall lower in the Synthetic conditions. } 
    \label{fig:result3} \vspace{-0.4cm}
\end{figure*}

\section{Computational Models}

Formally, the environment is defined by the set of agent states $S$, environment states $W$, the set of goals $G$, and the set of actions $A$. Let $s_t \in S$ be the agent’s state at time $t$, $w \in W$ be the world state, $g \in G$ the agent’s goal (a target), $a_t \in A$ the agent’s action at time $t,$ and $\tau$ the time of making the inference. By Bayes rule:
\begin{equation}
    \Pr(g|s_{1:\tau}, w)\propto \Pr(g|w)\Pr(s_{1:\tau}|g, w)
\end{equation}
Here $\Pr(g|w)$ is the prior over targets. We assume that all $18$ targets are equally likely, however in the general case the prior can be non-uniform, for example, it could reflect a preference to reach for closer targets. The goal of our model, and of heuristic controls, is 
to approximate $\Pr(s_{1:\tau}|g, w)$ under different computational assumptions.

\subsection{Generative Body Kinematics (BodyGen)}
The BodyGen model computes energy-minimizing probabilistic motion plans using an actor's body model. This forward planning procedure is then inverted to perform goal inference. 
We first create a skeletal model in MuJoCo \cite{todorov2012mujoco} that represents the body proportions of each actor. 
Second, we use a kinematic planning engine to generates possible reaching trajectories to all possible targets. Third, we calculate the probability of a given target based on the proximity of the generated trajectories to the observed actor's movement. The three steps are described in details below (see also Fig.\ref{fig:1}B). 

\textbf{Skeletal model construction.}
The skeletal models were built using the sizes, proportions and joint constraints aggregated over the motion captures of each actor. 
The skeletal models were driven by joint actuators with parameters for peak joint torque, joint damping, joint stiffness and torso moving speed. These values of there parameters were manually tuned during model design to produce realistic-looking movement.
Because of high computational costs of modeling walking, in the current work modeled only the actor's torso, as attached to a moving platform with constraints.
However, the computational principles of using inverse body kinematics for inference discussed in this work can apply to more general scenarios -- such as scenarios that include climbing, walking, and jumping, as well as other types of body models, such as models driven by muscle mechanics~\cite{jiang2018data, jiang2019synthesis}.
Other objects present in the scene in the experimental setup (e.g. table, chair, targets and obstacles), were reconstructed in MuJoCo using the recorded 3D proportions and sizes of these objects.

\textbf{Forward planning engine.}
We implemented a stochastic trajectory optimization with long-term value approximation similar to \cite{lowrey2018plan}, with the utility function is defined as $-p_1d+p_2h-p_3\mathbf{energy}-p_4{\mathbf{contact}}$, where $p_{1\cdots4}$ are free parameters of the skeletal model, $d$ is the distance between the wrist and target, and $h$ is the distance of the actor's head from the table surface. The model's movements were optimized to minimize energy expenditure, while penalizing contact force to avoid collisions. The $h$ parameter expresses a balancing constraint -- even though energy is often minimized more effectively by diving for the target, we enforce a constraint of reaching for the target while maintaining balance. We train a neural network to approximate the value function for each actor in each condition. The neural network approximately maps current human state with intended target to the expected value. We use an ensemble of 5 neural networks, each of which has 2 layers with 64 hidden parameters each. tanh is used as activation function. Training is performed on the random chosen human starting state and goal position. MPPI\cite{williams2017autonomous} algorithm is used to generate multiple reaching trajectories for local planning and iteratively improve the value function. We define $\text{Plan}(g, w, cs)$ as the function of generating future states given goal $g$ and starting state $cs$. The use of model predictive control (MPPI) in this case is mainly to reduce the unnecessary movements of planned motion. %

\textbf{Inverse planning as inference.}
To use the above system for inverse planning, let $q$ be the look-back history length. By assuming Markovian property:

\begin{equation}
\Pr(s_{1:\tau}|g, w)\propto \prod\limits_{t\in{2\cdots \tau }}\Pr(s_t|s_{t-1},g,w) 
\propto \prod\limits_{t\in{2, 2+q, 2+2q\cdots \tau}}\Pr(s_{t:t+q-1}|s_{t-1},g,w) 
\end{equation}

Since human movement and simulated movement are sampled from different continuous sequences, the two sequences need to be aligned using Dynamic Time Warping (DTW). We used DTW with a relatively long period $q$ to reduce the alignment noise, and estimated distance based on the wrist joint alone to reduce the number of free parameters. In a general case, this estimate can be refined by including other state information, such as body-mass centre, wrist velocity and acceleration of the wrist.
We run the planner multiple times to get the mean distances differences.
\begin{equation}
\Pr(s_{t:t+q-1}|s_{t-1},g,w)
\approx exp(-\beta_3\overline{\text{DTW}(\text{WP}(s_{t:t+q-1}),\text{WP}(\text{Plan}(g,w,s_{t-1})))})
\end{equation}

Here $\text{WP}(s)$ is the wrist position for state $s$.

Having defined our main BodyGen model, we next consider which simpler heuristics could be used to model energy minimization without inverse planning. Each of the heuristics described below assumes that the actor's goal is to minimize the distance between one of the actor's wrists and the desired target. Each heuristic makes different computational assumptions about the actor's constraints.

\subsection{Distance Heuristic} The Distance Heuristic sets the probability of each target as inversely proportional to the distance of the actor's wrist to the target, ignoring the history of the actor's movements. Denote $\text{TP}(g)$ as the target position of goal $g$, and $d$  the Euclidean distance function. The heuristic probability is defined as follows:
\begin{equation}
\Pr(g|s_{1:\tau}, w)\approx exp(-\theta d(\text{WP}(s_\tau),\text{TP}(g)))
\end{equation}

\subsection{Linear Extrapolation Heuristic (LinH)}
One simple way to extrapolate a reaching action is to assume that the agent intends to move to the target in a straight line. We extrapolate the actor's wrist position to a straight line, compute the shortest distance from each target to this line, and infer the target probability as inversely proportional to this distances. Formally, let $h_1$ be the look-back history length, $\alpha_1$ be the look-back probability length, $p(g)$ be the position of target $g$, let $l(s_{i:j})$ be the linear fit of endpoints from states $s_i\cdots s_j$ and let $d_{\text{linear}}(l, p)$ be the shortest Euclidean distance from point $p$ to line $l$. The heuristic probability is defined as: 
\begin{equation}
\Pr(s_{1:\tau}|g, w) \approx \prod_{k\in0,1\cdots\alpha_1}\exp(-\beta_1d_{\text{linear}}(l(s_{\tau - k-h_1 +1 :\tau-k}), p(g) ))
\end{equation}

\subsection{Parametric Curve Extrapolation Heuristic (ParamH)}
An important perceptual assumption in studies of biological motion, is that human and animal extremities move along circular trajectories~\cite{johansson1973visual}, since limbs are attached to a torso and have limited degrees of freedom. Thus, ParamH assumes a simple constraint on wrist motion, and extrapolates the endpoints of the actor's wrists to a parametric parabola curve $(a_1x^2+b_1x+c_1,a_2x^2+b_2x+c_2,a_3x^2+b_3x+c_3)$, to infer the target probability as inversely proportional to the shortest distance from the target to the curve. 
Let $pc(s_{i:j})$ be the parametric curve fit of endpoints from states $s_i\cdots s_j$ and let $d_{\text{pc}}(l, p)$ be the shortest Euclidean distance from point $p$ to line $l$. The probability is defined as: 
\begin{equation}
\Pr(s_{1:\tau}|g, w) \approx \prod_{k\in0,1\cdots\alpha_2} \exp(-\beta_2d_{\text{pc}}(pc(s_{\tau - k-h_2 +1 :\tau-k}), p(g) ))
\end{equation}

\subsection{Model Fitting}
We used Maximum Likelihood Estimation to fit the free parameters of each model using cross-validation. We divided all trials in each condition into two sets. Targets with odd indexes (i.e. 1,3,5..) were assigned to the training set, and the remaining targets comprised the test set. The same train-test split was used for all models. 
We used Nelder-Mead method to fit parameters ($q,\alpha_1,\alpha_2$), and Adam optimizer to fit the remaining parameters ($\beta_1,\beta_2,\beta_3,\theta$). 

\section{Behavioural experiments}

We conducted three behavioural experiments to measure human performance on the Target Reaching Task. If humans use inverse body kinematics to infer intentions of humans agents, then the BodyGen model should be better at predicting human inference compared to other models. However, since inverse body kinematics is computationally costly, people may use it only when necessary, and rely on the heuristics in easy scenarios (e.g. when an actor's hand is close to the target).

\begin{figure*}[t]
\centering
    \includegraphics[width=\textwidth]{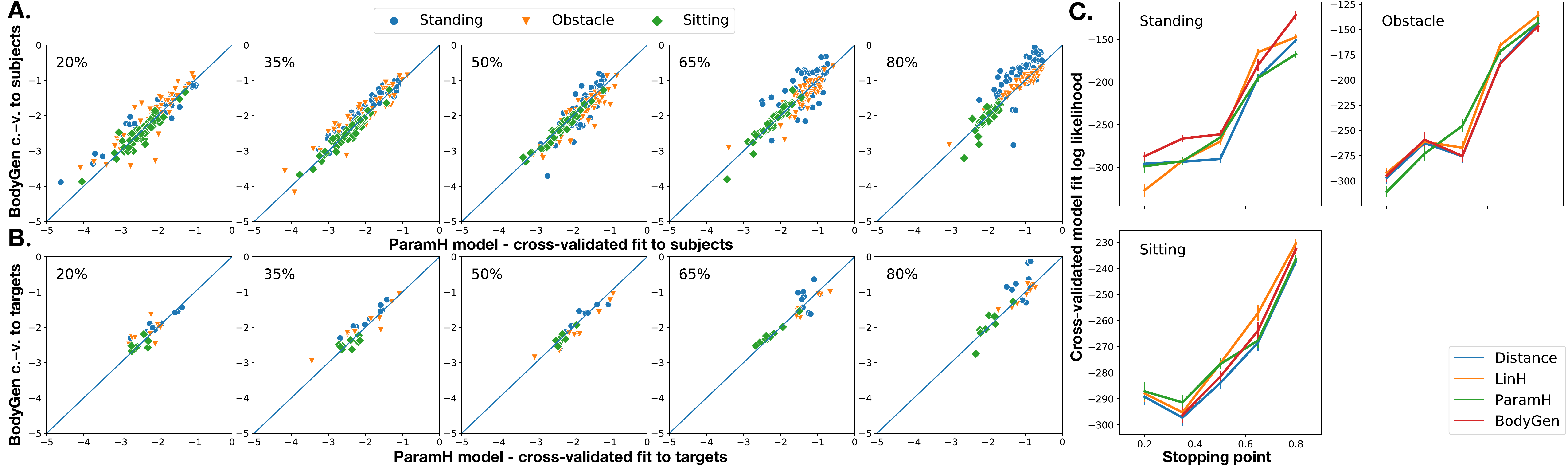}
    \caption{  Model comparison, Experiment 1. 
    \textbf{(A)} Comparing fit of the BodyGen and ParamH to subjects across stopping points. Each dot is a subject.
    \textbf{(B)}  Comparing fit of the BodyGen and ParamH to targets. Each dot is a target. %
    \textbf{(C)} Comparing cross-validated model fits shows that the BodyGen model has a slight advantage over other models in the Standing conditions at early stopping points.
    Error bars show 95\% confidence intervals.
    }
    \label{fig:result4} \vspace{-0.6cm}
\end{figure*}

\subsection{Experiment 1}
We evaluated human performance in 3 types of environments and two presentation styles: \textit{ \{Standing, Obstacle and Sitting\} } $\times $ \textit{ \{Face Blur Video, MuJoCo Synthethized\}} --  6 experimental conditions in total. All environments contained 18 targets, placed on a table in a 3x6 layout, as shown in Fig.~\ref{fig:1}C. 
In the Standing environment the actor stood behind the table, and could reach all potential targets without walking. In the Sitting environment the actor sat in a chair, placed at the right side of the table, and had to walk to reach objects on the left side of the table.
In the Obstacle environment a Plexiglas obstacle was placed in the middle of the table. The actor had to reach over the obstacle, or walk around it, to retrieve certain objects.  
Each environment was presented as a \textit{Face Blur}  -- a video recording with actor's face blurred out, and as a \textit{MuJoCo Synthethized} animation -- a reconstruction of the 3-D positions of the actor rendered on a stick-man figure. The animated stimuli provide human observers with the same input as the model, to control for the noise in the recorded 3D positions that may affect the model's performances.

\textbf{Subjects} We recruited 80 subjects (30 subjects for the three Face Blur condition, and 50 subjects for the three Synthesized conditions) on Amazon Mechanical Turk, who were paid for 50 minutes of work. This, and the following experiments, recorded no personally identifiable information, and were approved by our institutional IRB board. 

\textbf{Procedure} The experiment was presented in a web-browser, using an interface developed in our lab. After reading instructions and providing informed consent, subjects completed a session of 6 practice trials, in which they received feedback. 
On each trial a subject pressed a `Start' button to play a video. The video was paused at one of the stopping points ($20\%$, $35\%$, $50\%$, $65\%$ and $80\%$ of the total video duration), at which point the subject guessed the intended target by clicking on it. During the practice trials the subjects received an immediate feedback (either a big green check-mark, or a red cross, super-imposed over the video), after which they also saw the video play until the end.

The actual experiment did not include trial-based feedback. During the actual experiment the videos paused for the subject to respond, and after responding, the subject was immediately forwarded to the next trial. Subjects were given feedback on their aggregate accuracy after every 20 trials. The experiment lasted on average 45 minutes, subjects had an opportunity to take a break during this time. The subjects were told that we will reject submissions with poor performance, however no submissions were rejected. Detailed Experiment instructions and screenshots are given in the Supplement.

\textbf{Stimuli} The stimuli comprised 36 videos (18 targets, two actors) for each of the 6 conditions. Each stimulus was shown to each subject once, and was paused at a random at one of the 5 stopping points. 

\textbf{Empirical Results}
Fig.~\ref{fig:result3}A summarizes human performance across different targets in six conditions. For simple targets (e.g. target 3 in the center), people achieved $50\%$ accuracy after seeing as little as $50\%$ of the trajectory. In the Obstacle condition, the harder target (e.g. target 14 in middle of the bottom row) was almost never identified correctly. Fig.~\ref{fig:result3}B summarizes human performance aggregated across targets.  Overall, Sitting and Obstacles conditions were more challenging, however, for all conditions and targets, subjects' accuracy increased at later stopping points. Due to noise in the 3D recording the performance was lower in the Synthesized conditions.

\textbf{Model-based Results}
Fig.~\ref{fig:result4}C summarizes the model fit to individuals across stopping points and conditions. The BodyGen model had a slight overall advantage over heuristics in the Standing conditions at early stopping points.
Fig.~\ref{fig:result4}A. shows the models' fit to individual subjects, indicating that at each stopping point a fraction of individuals relied on BodyGen information for inference.  Comparing the models' performance on individual targets (see Fig.~\ref{fig:result4}B), shows that, for certain targets, the BodyGen model was better at predicting human inferences compared to the heuristics(For example, Target 5 and 8). 

\begin{figure*}[t]
\centering
    \includegraphics[width=0.95\textwidth]{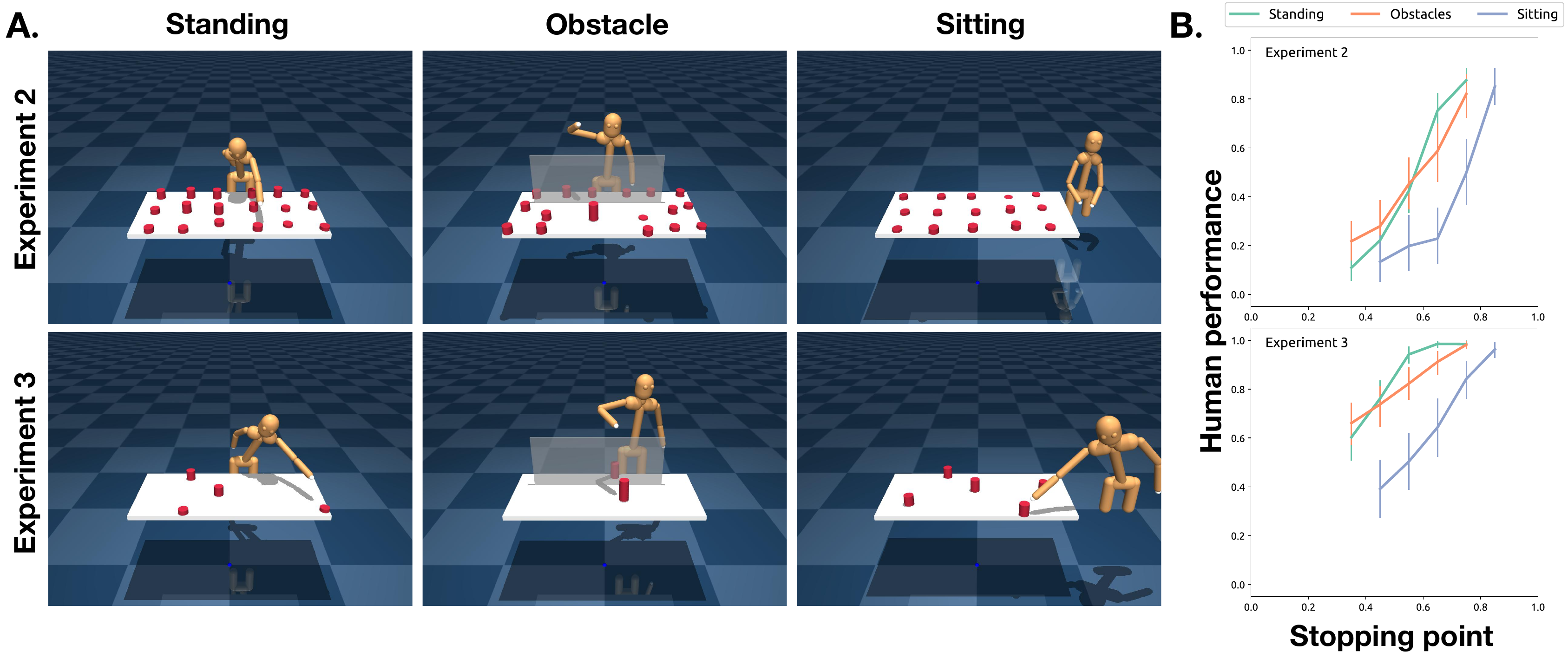}
    \caption{ { \small \textbf{A.} Screen-shots of stimuli used in Experiments 2 and 3.} \textbf{B.} Human performance in Experiments 2 and 3 in the three conditions aggregated across targets. Error bars indicate 95\% confidence intervals. }
    \label{fig:stim23} %
\end{figure*}

These results suggest that, at least for certain targets, and in certain scenarios, humans may be using body kinematics for inference. However, the noise in the 3D position recording, and the high number of distractors, may have reduced the discriminating power of our experiment, given that in the MuJoCo Synthesized conditions humans were unable to correctly identify certain targets even after the entire trajectory was shown. In the second experiments we aimed to partly control for the noise in 3D position, by adjusting the targets height and location on the table to match the average human wrist end-point, when actors reached for that target.

\begin{figure*}[t]
\centering
    \includegraphics[width=0.95\textwidth]{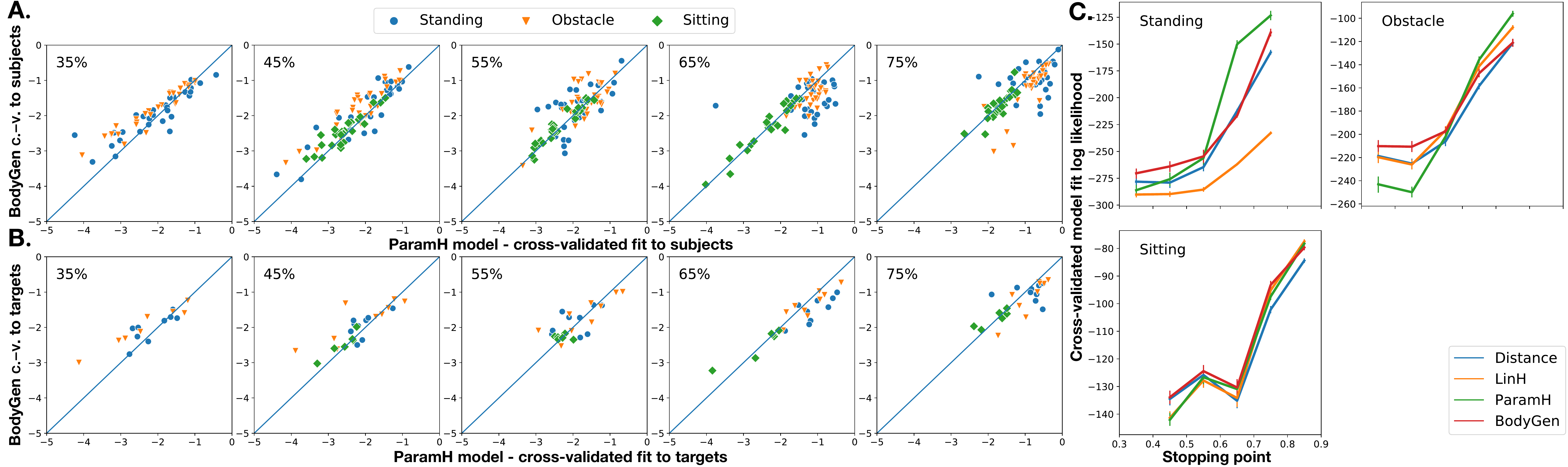}
    \caption{  Model comparison, Experiment 2. 
    \textbf{(A)} Comparing fit of the BodyGen and ParamH to individual subjects. %
    \textbf{(B)}  Comparing fit of the BodyGen and ParamH to targets, with model fit optimized across people. %
    \textbf{(C)} Comparing cross-validated model fits shows that the BodyGen model has a slight advantage over other models in the Standing and Obstacle conditions at early stopping points.
    Error bars show 95\% confidence intervals. 
       }
    \label{fig:e2} \vspace{-0.6cm}
\end{figure*}

\subsection{Experiment 2}

In the second experiment our goal was to remedy the noise in the 3D recording by shifting the targets to the ending position of the human wrist. We followed the procedure described in Experiment 1  with a different set of stimuli, all of which were synthesized animations made with MuJoCo. Each subject saw MuJoCo animated stimuli in three conditions (Sitting, Standing, Obstacle), which were presented in a randomized order, and paused at one of the following stopping points: $35\%, 45\%, 55\%, 65\%$ and $75\%$. The stopping points in the Sitting condition were shifted forward by $10\%$. There were $140$ trials in total. 

\textbf{Subjects} We recruited 30 subjects on Amazon Mechanical Turk, who were paid for 40 minutes of work. None were excluded.  

\textbf{Stimuli} The stimuli were created by modifying the three synthesized conditions (Sitting, Standing, Obstacle) of Experiment 1, by shifting the object positions to match the recorded position of the actors wrist at the end of the reaching action. If the target's adjusted position was outside the table surface, the target was removed, meaning that some stimuli in the second experiment had fewer than 18 targets. Examples of stimuli are shown in Fig.~\ref{fig:stim23}A (top).

\textbf{Empirical Results}
Fig.~\ref{fig:stim23}B summarizes human performance across conditions. Expected chance performance was around $0.06$. In the Standing and obstacle condition human performance was above chance at $45\%$ of the video, and in the sitting condition at $65\%$. Compared to the first experiment, humans were able to identify targets with higher accuracy, attesting to the success of our experimental manipulation.

\begin{figure*}[t]
\centering
    \includegraphics[width=0.95\textwidth]{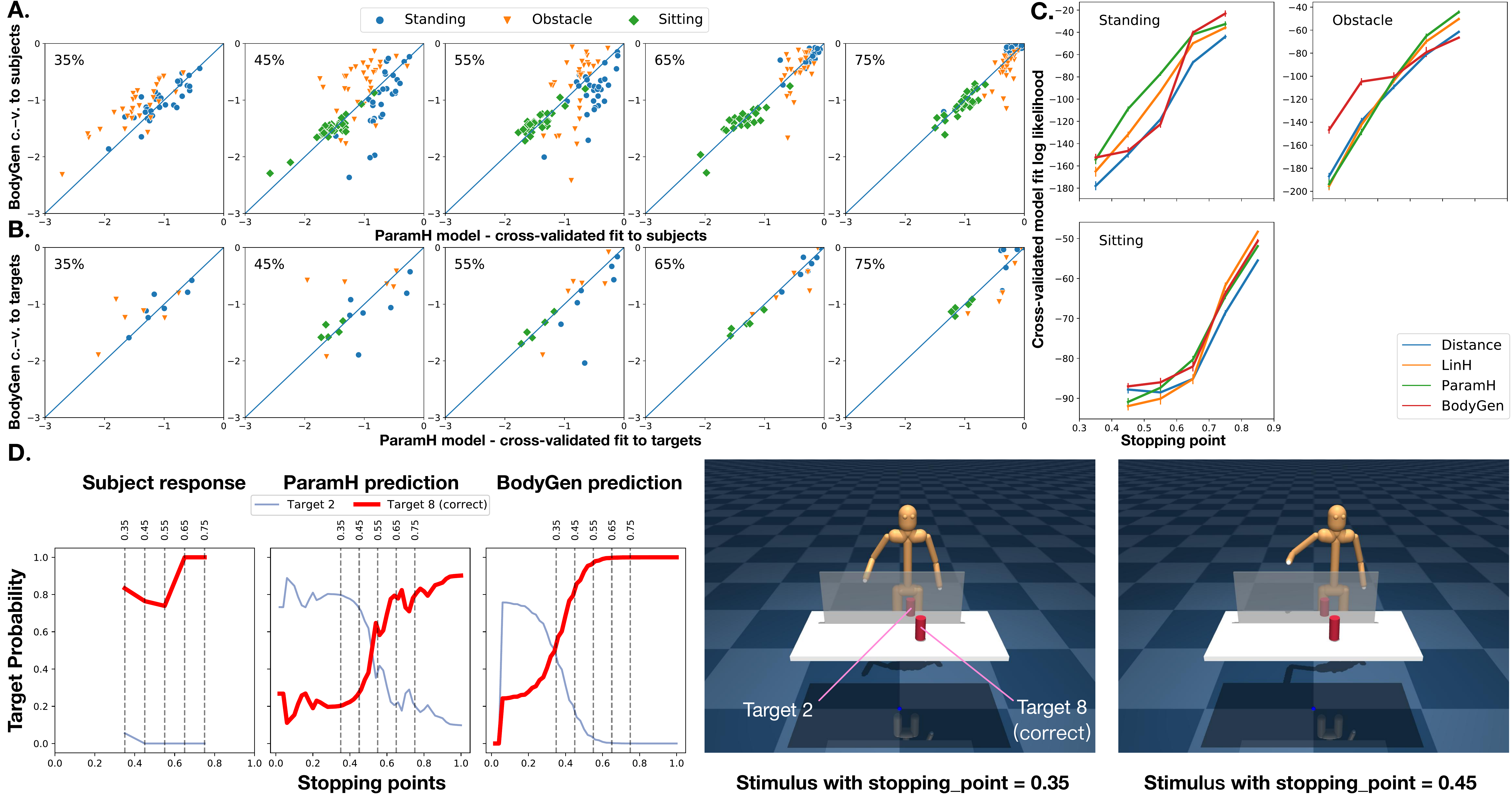}
    \caption{  Model comparison, Experiment 3. 
    \textbf{(A)} Comparing fit of the BodyGen and ParamH to subjects across stopping points and conditions. In the Obstacle condition during the $35\%$ and the $45\%$ stopping points most individuals were better predicted by the BodyGen model. 
    \textbf{(B)}  Comparing fit of the BodyGen and ParamH to targets. BodyGen was better at predicting human inferences about a number of targets during early stopping points. 
    \textbf{(C)} Comparing cross-validated model fits shows that the BodyGen model has an advantage over the heuristics in the Obstacle condition at early stopping points.
    Error bars show 95\% confidence intervals. \textbf{(D)} An example of a stimulus on which BodyGen outperforms ParamH. Target 8
    is correctly identified by the BodyGen model as soon as the robot starts to lift its arm to reach over the obstacle. ParamH is able to identify this target only after the actor's hand has passed over the obstacle, toward its other side.  }
    \label{fig:e3} \vspace{-0.6cm}
\end{figure*}

\textbf{Model-based Results}
Fig.~\ref{fig:e2}C summarizes the models' fit to individuals across stopping points and conditions. All models made increasingly acurate predictions over time. The BodyGen model had a slight advantage over other models in the Obstacle conditions at early stopping points, and the ParamH performed relatively better at late stopping points in the Standing condition.
Fig.~\ref{fig:e2}A. shows the models' fit to individuals aggregated over targets, indicating that at each stopping point a fraction of subjects relied on BodyGen information for inference, and this fraction was higher at earlier stopping points. 
Comparing the models' performance on individual targets, as shown in Fig.~\ref{fig:e2}B, revealed that for certain targets, the BodyGen model was better at predicting human inferences compared to ParamH, such as, for example, target 8 shown in~\ref{fig:e3}D. The relative fit of heuristics improved at later stopping points.

There results replicate the main findings of Experiment 1, showing shown that the relative reliance on body kinematic information is stronger in the beginning of the trajectory in environments with obstacles. However, this advantage disappears in the later stages of the reaching actions, where subject increasingly shift toward heuristics, possibly to minimize mental effort and decision time.
At the same time, the differences between our model and heuristics are still quite small. We reasoned that this may be due to the high number of distractors, which make it hard to make an accurate inference early, when the benefit of using the BodyGen model is the strongest. In the next experiment we test this possibility by reducing the number of distractors.

\subsection{Experiment 3}

In the third experiment we used the target positions computed in Experiment 2, while using a smaller number of objects of each trial. For each target, the distractors were chosen by randomly sampling 2 to 6 non-adjacent distractors. The experimental procedure was otherwise identical to the procedure used in Experiments 1 and 2. 

\textbf{Subjects} We recruited 30 subjects on Amazon Mechanical Turk, who were paid for 40 minutes of work. None were excluded. 

\textbf{Stimuli} Each subject saw stimuli in all three conditions (Standing, Sitting, Obstacle), which were presented in a randomized order, and paused at one of the following stopping points: $35\%, 45\%, 55\%, 65\%$ and $75\%$. The stopping points in the Sitting condition were shifted forward by $10\%$. There were $160$ trials in total.  

\textbf{Empirical Results}
Fig.~\ref{fig:stim23}B summarizes human performance across conditions. Expected chance performance was between $0.2$ and $0.5$ depending on number of possible targets. Human performance was above chance at all stopping points and in all conditions. 

\textbf{Model-based Results}
Fig.~\ref{fig:e3}C summarizes the models' fit to individuals across stopping points and conditions. All models made increasingly accurate predictions over time. The BodyGen model had a advantage over the heuristics in the Obstacle conditions at early stopping points.
Fig.~\ref{fig:e3}A. shows the models' fit to subjects, indicating that in the Obstacle condition, at each stopping point a fraction of individuals relied on kinematic body simulation approximated by BodyGen. 
Comparing the models' performance on individual targets, as shown in Fig.~\ref{fig:e3}B, revealed that for certain targets the BodyGen model was better at predicting human inferences compared to the ParamH during early stopping points. Figure~\ref{fig:e3}D shows an example of a stimulus on which BodyGen outperforms ParamH during the early stage of the reaching action. Like humans, BodyGen infers the correct target from seeing the actor begin to reach over the barrier. In contrast, ParamH is unable to make a correct prediction until after the actor's wrist has passed the high point above the barrier, which allows the model to complete the parabolic extrapolation.

\section{General Discussion}

\textbf{Contributions} We have developed an inverse kinematic model of human intention inference, and evaluated its performance against three heuristics in three behavioural experiments using a novel Target Reaching Task. We found that in simple cases (e.g. the actor is close to touching the target) simple extrapolations of wrist trajectory can accurately predict human intention inferences. However, in certain more complex scenarios -- notably when an agent needs to reach over an obstacle, explaining human intention inference relies on inverse body kinematics. We show that body kinematics is particularly important during the early stages of reaching actions, as well as in environments with obstacles, where the human agent's trajectory can not be extrapolated in a trivial way. Understanding how humans interpret each others' motor trajectories has  multidisciplinary benefits to robotics (better collaboration), AI (modeling planning), and cognitive science (theory of mind). 

\textbf{Future work} In future work we intend to investigate the consistency in the use of goal inference strategies between individuals, and between tasks. For example, we intend to test whether people revert to heuristics to minimize cognitive effort under cognitive load, even when the use of heuristic decreases accuracy. We also intend to investigate whether people rely on the best-performing heuristic in a given type environment, and when do they rely on the more costly inference based on inverse body kinematics. Future work should also investigate intention inferences in other complex tasks, as well as complex task-and-motion scenarios involving sequences of actions.  

\textbf{Limitations} 
One major limitation of the current research is the high noise in the 3D recording of the objects and actor body positions. As we discovered during data post-processing, actors have sometimes interpreted the instructions to reach an object differently --- as either as `holding the object in hand' or `touching the object with finger-tips', which resulted in inter-trial noise in how closely the wrist approached the target.
Further, when the intended object is hard to reach, a human may stretch, stand on toe-tips, and so on, to extend the range of motion -- such actions are natural to humans, but can not be represented under the current model. The BodyGen model illustrates the use of inverse body kinematics in a proof-of-concept way, however its performance as compared against heuristics is likely a lower-bound, due to modelling only a few aspects of how bodies can move. The BodyGen model can be improved by using more elaborate body models, for example, by modelling bodies as driven by muscles~\cite{jiang2018data, jiang2019synthesis} and by exploring other algorithms of controlling them (e.g. \cite{peng2021amp}).
Further, it is unclear to what extent our current skeletal body model used for inverse kinematics is a reasonable approximation of the internal models of body movement that humans actually have. Future work should evaluate several classes of body models, in order to find to what extent the accuracy of BodyGen inference depends on the specific implementation of body kinematics.

\textbf{Potential negative impact} The long-term goal of our work is to reverse-engineer how humans think. This line of research is still in its infancy. However it could enable future research with potentially harmful consequences. For example, algorithms could exploit the sub-optimalities in human intuitions to adapt to human decision-making in order to control it. It also raises the possibility of engineering an AI that interprets how humans plan, and uses this inference to anticipate and prevent human actions. %

\printbibliography
\end{document}